%% file: dgplvm.tex
\documentclass{article} 
\usepackage{iclr2016_workshop,times}
\usepackage{hyperref}
\usepackage{url}

\usepackage{caption}
\usepackage{subcaption}

\usepackage{graphicx} 
\DeclareGraphicsExtensions{.pdf,.jpeg,.png}

\usepackage{amsmath}
\usepackage{amssymb}

\input{notationDef.tex}

\input{definitions.tex}

\global\long\def\covarianceScalar{k}

\newcommand{\K}{\covarianceMatrix}

\newcommand{\zV}{\inducingInputVector}

\newcommand{\xM}{\inputMatrix}
\newcommand{\xV}{\inputVector}

\newcommand{\I}{\identityMatrix}
\newcommand{\yM}{\dataMatrix}
\newcommand{\yV}{\dataVector}

\newcommand{\cV}{\mathbf{c}}

\title{Unsupervised Learning with Imbalanced Data via Structure Consolidation Latent Variable Model}

\author{Fariba Yousefi$^1$, Zhenwen Dai$^2$, Carl Henrik Ek$^3$, Neil Lawrence$^4$ \\
$^{1,4}$ University of Sheffield, Sheffield, UK \\
$^2$ Inferentia Ltd, UK\\
$^3$ University of Bristol, Bristol, UK \\
$^{1,4}$ \texttt{\{f.yousefi, n.lawrence\}@sheffield.ac.uk} \\
$^2$ \texttt{zhenwendai@gmail.com} \\
$^3$ \texttt{carlhenrik.ek@bristol.ac.uk} \\
}

\begin{document}

\maketitle

\begin{abstract}
Unsupervised learning on imbalanced data is challenging because, when given imbalanced data, current model is often dominated by the major category and ignores the categories with small amount of data. We develop a latent variable model that can cope with imbalanced data by dividing the latent space into a shared space and a private space. Based on Gaussian Process Latent Variable Models, we propose a new kernel formulation that enables the separation of latent space and derives an efficient variational inference method. The performance of our model is demonstrated with an imbalanced medical image dataset.
\end{abstract}

\section{Introduction}

In many medical applications, e.g. pathology, negatively labelled data is extremely easy to obtain (e.g. healthy cells). Positive labels, on the other hand, can be harder to acquire (e.g. particular disease morphologies). These massively unbalanced problems are challenging for most algorithms because the negative class tends to dominate the objective function and the resulting model performs poorly. In practice it is often better to throw away much of the negative data and rebalance the data set. 

Unsupervised learning has been attracting a lot of attention as it has the potential to serve as an underpinning technology for a range of challenges such as generative modeling, missing data imputation and coping with multiple data modalities. Unsupervised learning can also be applied to a wider range of data sets, because it does not rely on having carefully labelled data available.

In this paper we explore the possibility of using a variant unsupervised learning algorithm to solve the problem of label balance. We build latent variable models that can simultaneously accommodate a very large number of negative examples, sharing their characteristics appropriately with the positive class, while simultaneously allowing the model to characterise the manner in which the positive class is differently characterised through preserved (or private) latent spaces that are separately learned for each class. The resulting model does not suffer from the standard challenges in this domain. We compare with a variant of the discriminative GP-LVM (the model that underpinned GaussianFace) and show signficantly improved performance.

Our probabilistic latent variable model divides its latent space into a shared space of all the categories and a private space for each category \citep{Damianou:manifold12}. The shared space accounts for capturing the common regularities among categories (e.g. positive and negative class) and the private space is dedicated to model the variance specific to individual categories. Because the modelling of the private space is category specific, there is no domination of it's characteristics by the larger category. Thus the data in each category can be modeled appropriately while the common regularities are still exploited.

We implement the idea of shared and private space in the framework of Gaussian Process Latent Variable Models \citep[GPLVM,][]{Lawrence:pnpca05} by deriving a particular covariance function (kernel) that enables such separation. We exploit closed form variational lower bounds of the log marginal likelihood of the proposed model, which to provide an efficient approximation inference method.

The performance of our model is evaluated with a real image dataset, in which the positive and negative data are extremely imbalanced. We show our model still can learn from imbalanced data and perform well in both generative and discriminative tasks.

\section{Structure Consolidation Latent Variable Model}

We assume the dataset is represented as a set of fixed length vectors $\yM \in \Re^{N \times D}$, where $N$ is the number of data points and $D$ is the dimensionality of individual data points. Additionally, a label of category is associated with each data point, $\cV = (c^{(1)},\ldots,c^{(N)}), c^{(n)} \in \{1,\ldots,C\}$, where $C$ indicates the number of categories in the dataset. We aim at building a probabilistic model $p(\yM)$ that is robust when the numbers of data in different categories are highly imbalanced.

We assume the data associated with a set of latent representations $\xM \in \Re^{N \times Q}$, where $Q$ is the dimensionality of the latent space. The latent representations are related to the observed data through an unknown mapping function $f$ and $f$ follows a prior distribution that is defined as a Gaussian process,
\begin{equation}
\yV = f(\xV) + \epsilon, \quad f \sim \mathcal{GP}(0, k),
\end{equation}
where $\epsilon \in \mathcal{N}(0,\sigma^2\I)$ denotes the observation noise and $k$ is the kernel function. Given the observed data $\yM$, we wish to obtain a posterior estimate for both the latent representation $\xM$ and the unknown mapping function $f(\cdot)$.
In our model we separate the latent space into a shared space with the dimensionality $Q_s$ and a private space with the dimensionality $Q_p$. Therefore, a latent representation can be denoted as $\xV = [ \xV_s^\top, \xV_p^\top]^\top, \xV_s \in \Re^{Q_s}, \xV_p \in \Re^{Q_p}$, where $\xV_s$ and $\xV_p$ are the latent representations in shared and private space respectively. With the separated latent representation, we define the kernel function in our model as
\begin{equation}
k((\xV, c_{\xV}), (\xV', c_{\xV'})) = k_s(\xV_s, \xV_s') + k_p((\xV_p, c_{\xV}), (\xV_p', c_{\xV'})),
\end{equation}
where $k_s$ is the kernel function for the shared space and $k_p$ is the kernel function for the private space. The shared kernel can be any kernel function built on a vector space from the literature. However, the private kernel is defined to take the following form:
\begin{equation}
k_p((\xV_p, c_{\xV}), (\xV_p', c_{\xV'})) = 
\begin{cases}
k'(\xV_p, \xV_p'), & c_{\xV} = c_{\xV'}, \\
0, & c_{\xV} \neq c_{\xV'},
\end{cases}
\end{equation}
where $k'$ is the kernel function chosen to calculate the covariance and $c_{\xV}$ is the label of category for the data point $\xV$. We give a unit Gaussian prior distribution to latent representations $\xV \sim \mathcal{N}(0,\I)$. The log marginal likelihood for the proposed model can be derived as $\log p(\yM | \cV) = \log \int p(\yM | \xM, \cV) p(\xM) \dif{\xM}$.
There is no analytical solution for this marginal likelihood. We apply variational inference and derive a closed form lower bound of the log marginal likelihood, by following a sparse Gaussian process approximation \citep{Titsias:bayesGPLVM10}:
\begin{align}\small
\log p(\yM) \geq &\sum_{d=1}^{D} \tilde{F}_d(q) - \text{KL}(q(\xM)\|p(\xM)),\\
\tilde{F}_d(q) =& \log \Big[\frac{|\K_{uu}|^{\frac{1}{2}}}{(2\pi \sigma^2)^{\frac{N}{2}} | \beta \Psi_2 + \K_{uu}|^{\frac{1}{2}}} e^{-\frac{1}{2}\yV_{d}^\top W \yV_{d}} \Big]  -\frac{\psi_0}{2\sigma^2} + \frac{1}{2\sigma^2}\text{Tr}(\K_{uu}^{-1}\Psi_2),\label{equ:Ft_d}
\end{align}
where $W= \frac{1}{\sigma^2} \I - \frac{1}{\sigma^4} \Psi_1 (\frac{1}{\sigma^2}  \Psi_2+\K_{uu})^{-1}\Psi_1^\top$, and $\psi_0$, $\Psi_1$, $\Psi_2$ are the expectation of covariance matrices w.r.t.\ the variational posterior $q(\xM)$. In our model, these expectation are derived as
\begin{align}\small
\psi_0 
=& \sum_{n=1}^N \expectationDist{k_s(\xV^{(n)}_s, \xV^{(n)}_s)}{q(\xV^{(n)}_s)} + \expectationDist{k_p((\xV^{(n)}_p, \cV^{(n)}_{\xV}), (\xV^{(n)}_p, \cV^{(n)}_{\xV}))}{q(\xV^{(n)}_p)}\\
(\Psi_1)_{nm} 
 =& \expectationDist{k_s(\xV^{(n)}_s, \zV^{(m)}_s)}{q(\xV^{(n)}_s)} + \expectationDist{k_p((\xV^{(n)}_p, \cV^{(n)}_\xV), (\zV^{(m)}_p, \cV^{(m)}_\zV))}{q(\xV^{(n)}_p)} \\
 (\Psi_2)_{mm'} 
 =& \sum_{n=1}^N \expectationDist{k_s(\xV^{(n)}_s, \zV^{(m)}_s)k_s(\xV^{(n)}_s, \zV^{(m')}_s)}{q(\xV^{(n)}_s)} \nonumber \\
 &+ \expectationDist{k_p((\xV^{(n)}_p, \cV^{(n)}_\xV),(\zV^{(m)}_p, \cV^{(m)}_\zV)) k_p((\xV^{(n)}_p,\cV^{(n)}_\xV), (\zV^{(m')}_p, \cV^{(m')}_\zV))}{q(\xV^{(n)}_p)} \nonumber\\
 &+ \expectationDist{k_s(\xV^{(n)}_s, \zV^{(m)}_s)}{q(\xV^{(n)}_s)}\expectationDist{k_p((\xV^{(n)}_p, \cV^{(n)}_\xV), (\zV^{(m')}_p, \cV^{(m')}_\zV))}{q(\xV^{(n)}_p)} \nonumber\\
 &+ \expectationDist{k_p((\xV^{(n)}_p, \cV^{(n)}_\xV), (\zV^{(m)}_p, \cV^{(m)}_\zV))}{q(\xV^{(n)}_p)}\expectationDist{k_s(\xV^{(n)}_s, \zV^{(m')}_s)}{q(\xV^{(n)}_s)}
\end{align}
where $\zV$ and $c_z$ are the variational parameters known as inducing inputs and inducing labels.

%

\section{Experiment}

\begin{figure}[t!]
\centering
    \begin{subfigure}[b]{0.35\textwidth}
        \includegraphics[width=1.\linewidth]{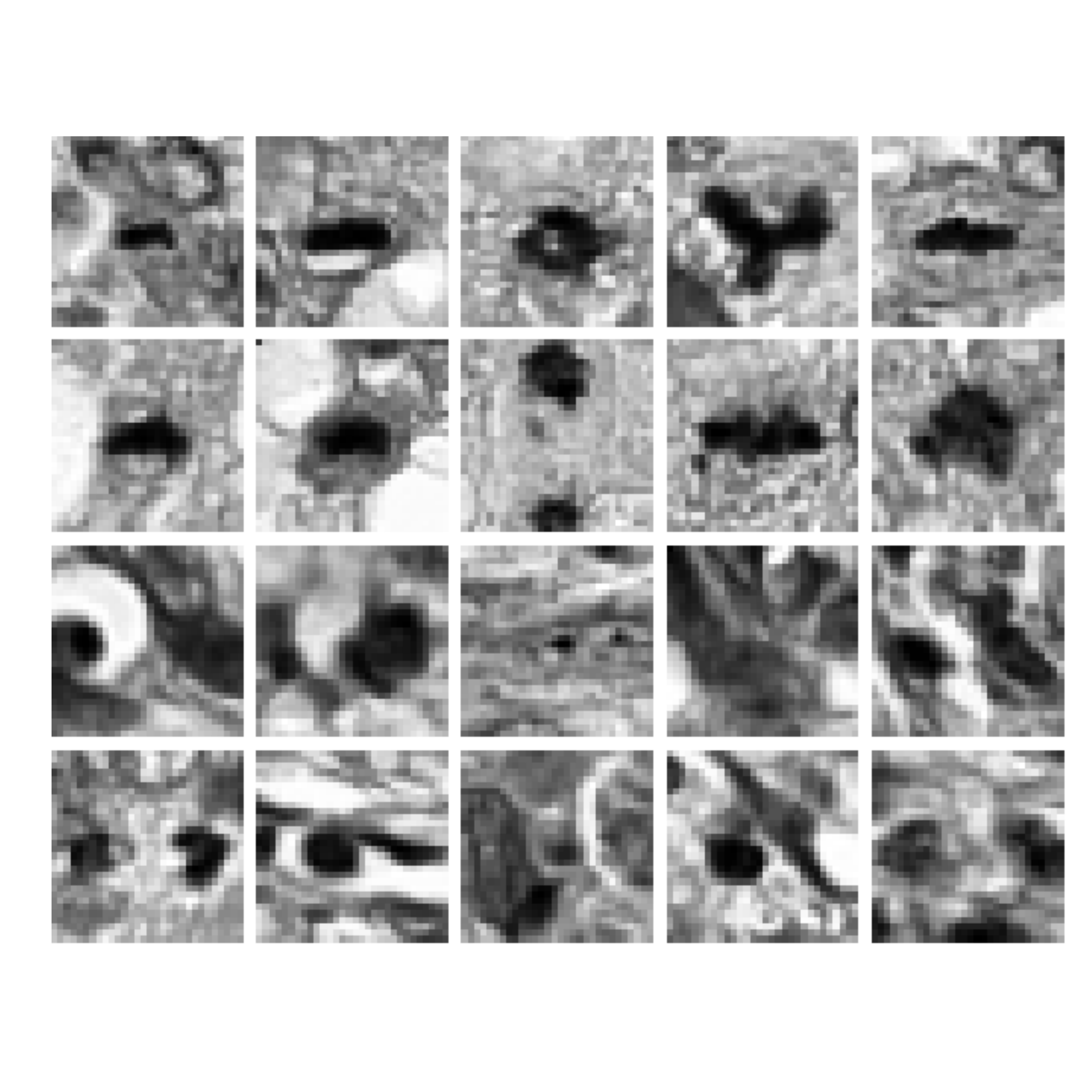}
        \caption{}\label{fig:data_samples}
    \end{subfigure}
    \begin{subfigure}[b]{0.35\textwidth}
    \includegraphics[width=1.\linewidth]{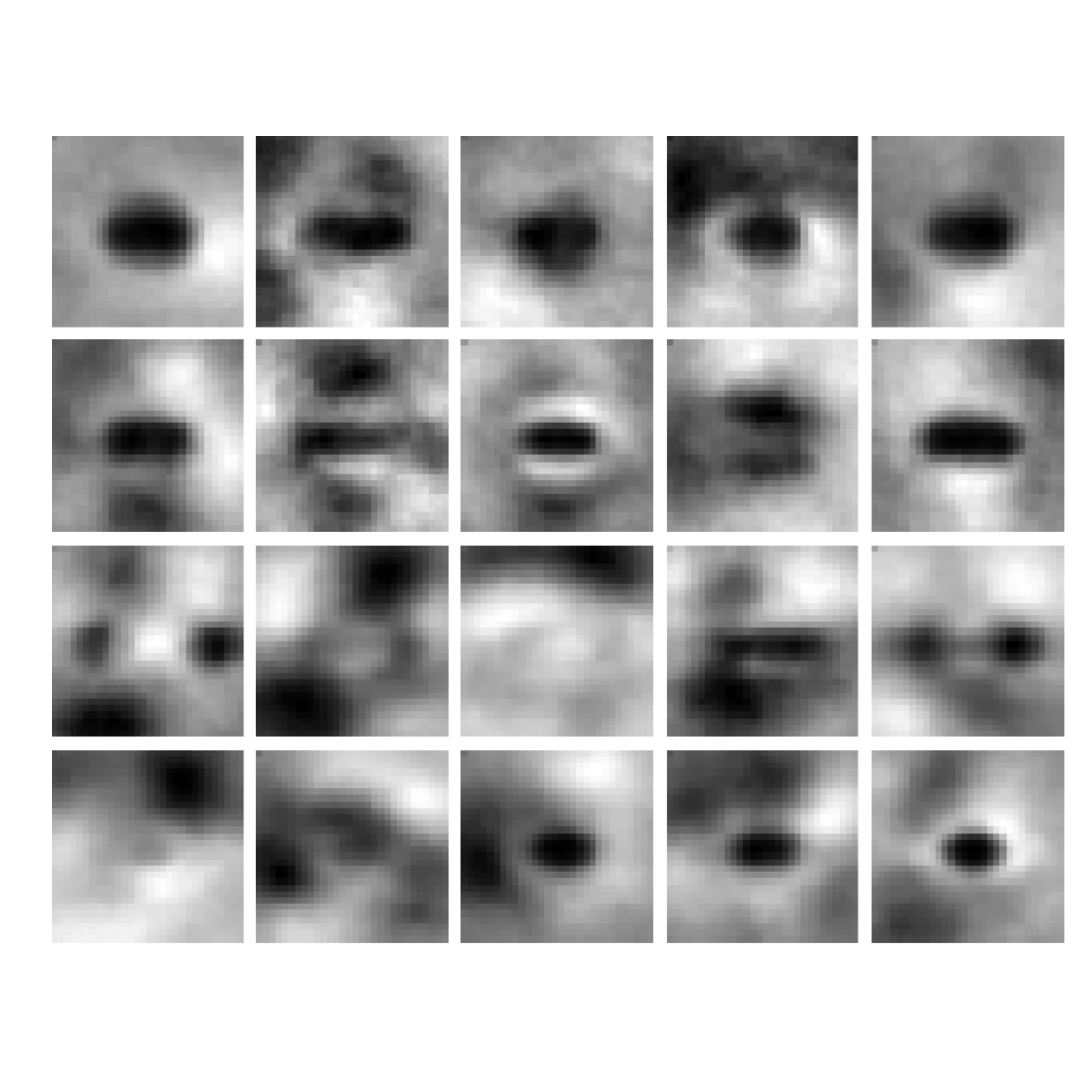}
    \caption{}\label{fig:gen_samples}
    \end{subfigure}
 \caption{\small(a) Some examples in the data sets. (b) Samples generated from the trained SCLVM. In both figures, the first two rows correspond to positive images and the last two rows correspond to negative images.}
\end{figure}

Mitosis detection is a stage in tumour assessment that involves determining whether individual cells are in mitosis (dividing to reproduce). These cells are rare. We used data from the assessment of mitosis detection algorithm 2013 \citep[AMIDA13,][]{veta2015assessment} challenge which is publicly available. The main goal of the challenge is to find proper mitosis detection methods that can be automatic or semi-automatic. We use the training set from the challenge, which consists of tissue images of $12$ patients and are annotated by human experts. We preprocess the tissue images with the algorithm by \cite{Snell:cell} and focus on the generated candidate image patches. The resulting image set contains 146,562 grey-scale image patches ($30 \times 30$ pixels), of which 550 are positive (mitosis) according to manual annotation.
We randomly take 80\% of positive images and 5,000 negative images as the training data. This gives in total 5,440 images. Some examples of the training data is shown in Figure \ref{fig:data_samples}.  We applied SCLVM to this dataset and used an exponentiated quadratic kernel for both the shared and private space and set the dimensionality of both shared and private space to five. 

\begin{figure}[t!]
\centering
\includegraphics[width=.745\linewidth]{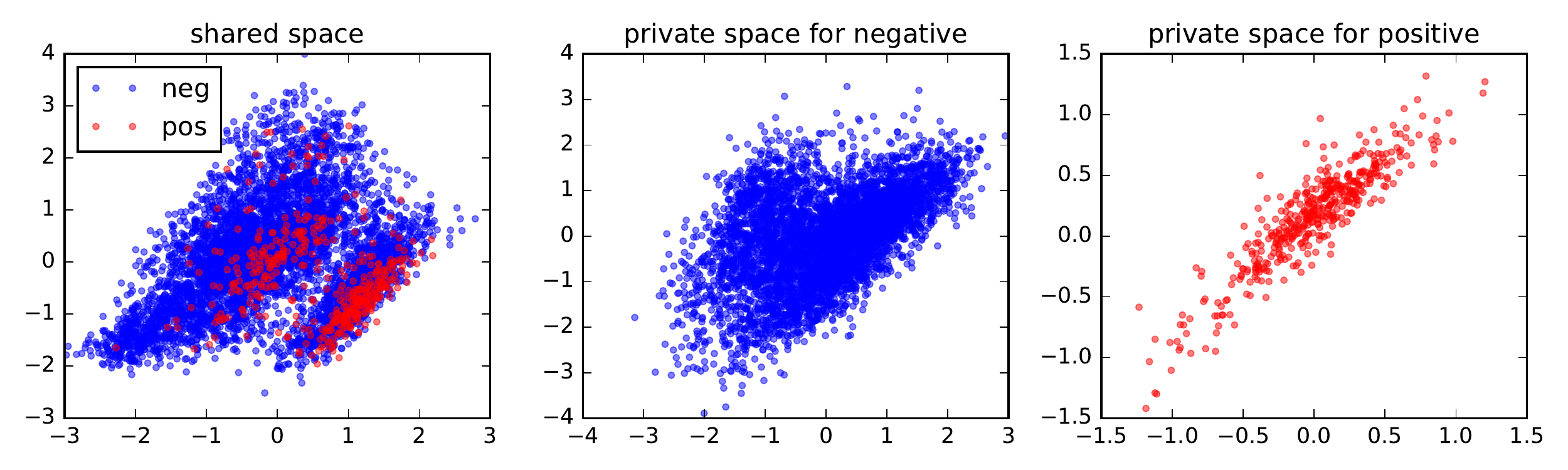}
\includegraphics[width=.245\linewidth]{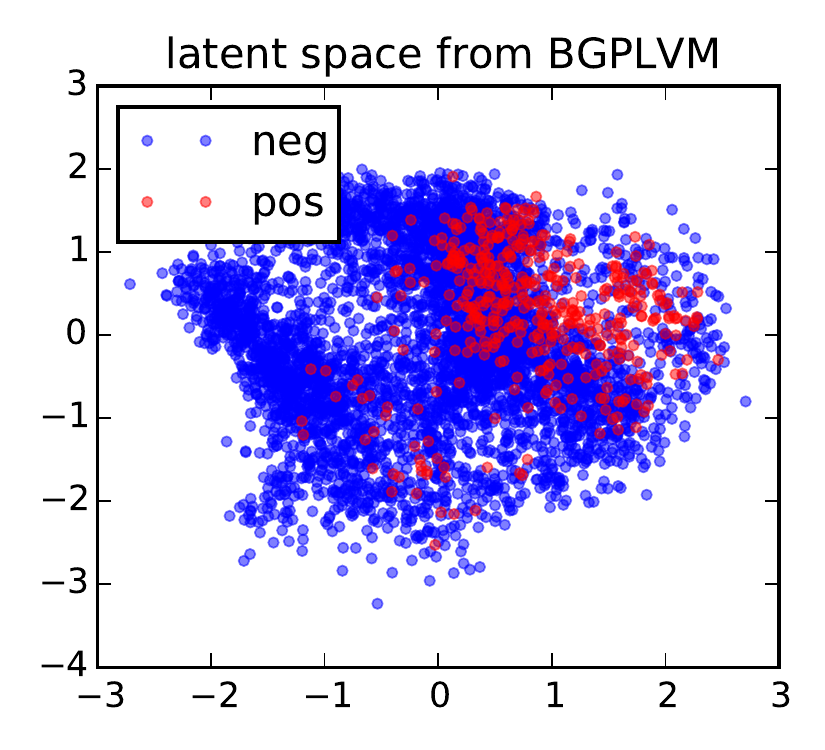}
\caption{\small The visualization of the training data in learned latent spaces. The first figure shows the positive and negative data in two of the shared dimensions. The second and third figures show the two of the private dimensions for the negative and positive data respectively. The fourth figure shows the learned latent space from BGPLVM.}\label{fig:latent_space}
\end{figure}

Both the latent representations and kernel parameters are optimized until convergence. The resulting latent space is visualized in Figure \ref{fig:latent_space}. The positive and negative images present similar structures in the shared space, which demonstrates the discovered common regularities, and their private space are significantly different from each other. To demonstrate the ability of SCLVM in balancing the modeling capabilities between imbalanced categories, we draw samples from the learned latent space of SCLVM for both positive and negative categories (see Figure \ref{fig:gen_samples}). The generated samples from positive and negative categories are clearly different from each other and they capture some characteristics of their own categories. We further evaluate the learned latent space by performing classification on test set (the rest of positive examples plus randomly sampled negative examples, in total 1000 images). We compare SCLVM with BGPLVM \citep{Titsias:bayesGPLVM10} and DGPLVM\footnote{Due to its complexity $O(N^3)$, we only use 1000 images for training (440 positive and 560 negative).} \citep{urtasun2007discriminative} with ten test sets. We apply weighted SVM with an exponentiated quadratic kernel on the latent space from the BGPLVM and DGPLVM. The results are shown in Table \ref{tab:classification}. Note that BGPLVM requires an additional classification model to be learned. This does not provide probabilities over the classes other than in the ad-hoc manner that an SVM will. Similarly DGPLVM does learn a space that reflects the class information, but it does not provide means to get the posterior over the classes. Our model is the only one that learns the classification jointly with the model and provides a principled way of getting probabilities over the classes.

\begin{table}[h!]
\begin{center}
\caption{Classification performance. The mean and standard deviation from ten test sets are shown.}\label{tab:classification}
\begin{tabular}{ c|c|c|c } 
& SCLVM & BGPLVM (SVM) & DGPLVM (SVM) \\
 \hline \hline
 precision & $0.426 \pm 0.024$ & $0.306 \pm 0.013$& $0.242 \pm 0.008$\\
 recall & $0.555 \pm 0.007$ & $0.827 \pm 0.000$ & $1.000 \pm 0.000$\\
 F1 score & $0.482 \pm 0.015$ & $0.447 \pm 0.014$ & $0.390 \pm 0.011$\\
\end{tabular}
\end{center}
\end{table}

\section{Conclusion}

We presented a probabilistic latent variable model that can cope with imbalanced data. We developed a kernel that separates the latent space into a shared spare and a private space. An efficient variational inference method is proposed by deriving a closed form lower bound of marginal likelihood. Beyond the shown example, the ability of jointly modelling multiple data categories and handling imbalanced datasets can be linked to many other areas such as transfer learning.

{\small
\bibliography{lawrence,other,zbooks,dgplvm}
\bibliographystyle{iclr2016_workshop}
}

\end{document}

%% file: notationDef.tex
\global\long\def\dataScalar{y}
\global\long\def\dataVector{\mathbf{\dataScalar}}
\global\long\def\dataMatrix{\mathbf{\MakeUppercase{\dataScalar}}}

\global\long\def\inputScalar{x}
\global\long\def\inputMatrix{{\bf \MakeUppercase{\inputScalar}}}
\global\long\def\inputVector{{\bf \inputScalar}}

\global\long\def\covarianceScalar{c}
\global\long\def\covarianceMatrix{\mathbf{\MakeUppercase{\covarianceScalar}}}

\global\long\def\inducingInputScalar{z}
\global\long\def\inducingInputVector{\mathbf{\inducingInputScalar}}

\global\long\def\eye{\mathbf{I}}
\global\long\def\identityMatrix{\eye}

\global\long\def\expectationDist#1#2{\left\langle #1 \right\rangle _{#2}}


%% file: definitions.tex
\usepackage{color}
\usepackage{verbatim}

\definecolor{brown}{rgb}{0.9,0.59,0.078}
\definecolor{ironsulf}{rgb}{0,0.7,.5}
\definecolor{lightpurple}{rgb}{0.156,0,0.245}

\ifdefined\blackBackground
\definecolor{colorOne}{rgb}{0, 1, 1}
\definecolor{colorTwo}{rgb}{1, 0, 1}
\definecolor{colorThree}{rgb}{1, 1, 0}
\definecolor{colorTwoThree}{rgb}{1, 0, 0}
\definecolor{colorOneThree}{rgb}{0, 1, 0}
\definecolor{colorOneTwo}{rgb}{0, 0, 1}
\else
\definecolor{colorOne}{rgb}{1, 0, 0}
\definecolor{colorTwo}{rgb}{0, 1, 0}
\definecolor{colorThree}{rgb}{0, 0, 1}
\definecolor{colorTwoThree}{rgb}{0, 1, 1}
\definecolor{colorOneThree}{rgb}{1, 0, 1}
\definecolor{colorOneTwo}{rgb}{1, 1, 0}
\fi

\ifdefined\blackBackground

\else

\fi

\newcommand{\dif}[1]{\text{d}#1}

\global\long\def\eye{\mathbf{I}}

\global\long\def\cut#1{}

\global\long\def\detail#1{}

